# LLMs as Deceptive Agents: How Role-Based Prompting Induces Semantic Ambiguity in Puzzle Tasks

**Seunghyun Yoo[*], Taeyoung Kim, Sihyun Yoo**
Carter Student AI Research (CSAI) Lab
{seunghyun050616, ttang1106t, sihyun20070801}@gmail.com

## Abstract

Recent advancements in Large Language Models (LLMs) have not only showcased impressive creative capabilities but also revealed emerging agentic behaviors that exploit linguistic ambiguity in adversarial settings. In this study, we investigate how an LLM, acting as an autonomous agent, leverages semantic ambiguity to generate deceptive puzzles that mislead and challenge human users. Inspired by the popular puzzle game "Connections," we systematically compare puzzles produced through zero-shot prompting, role-injected adversarial prompts, and human-crafted examples, with an emphasis on understanding the underlying agent decision-making processes. Employing computational analyses with HateBERT to quantify semantic ambiguity, alongside subjective human evaluations, we demonstrate that explicit adversarial agent behaviors significantly heighten semantic ambiguity—thereby increasing cognitive load and reducing fairness in puzzle solving. These findings provide critical insights into the emergent agentic qualities of LLMs and underscore important ethical considerations for evaluating and safely deploying autonomous language systems in both educational technologies and entertainment.

## 1 Introduction

The remarkable capabilities of contemporary Large Language Models (LLMs) have extended into domains that require sophisticated agentic decision-making and nuanced linguistic manipulation (Franceschelli & Musolesi, 2024). While AI creativity has traditionally emphasized novelty, value, and surprise (Colton, 2008), recent research has highlighted a fourth dimension: deceptiveness, defined as the intentional manipulation of ambiguity to mislead users (Wang et al., 2024). This agentic behavior—wherein the model actively exploits linguistic ambiguity—challenges conventional approaches to agent quality evaluation and raises critical issues regarding safety and fairness.

Linguistic ambiguity, arising from polysemy and semantic overlap, plays a pivotal role in cognitive processing and task complexity (Liu et al., 2023). According to Cognitive Load Theory (CLT), increased ambiguity imposes higher cognitive demands, directly impairing performance by prolonging response times and increasing error rates (Fox & Rey, 2024). Although previous studies have focused on embedding-based semantic analyses for detecting ambiguity (Mesgar & Strube, 2016), the explicit evaluation of adversarial agent behaviors remains relatively under-explored.

Motivated by the puzzle game "Connections" from The New York Times, our study compares puzzles generated by LLMs under different prompting conditions—namely zero-shot and role-injected adversarial prompts—against human-crafted counterparts. By utilizing HateBERT for computational evaluation and integrating subjective human assessments, we examine how adversarial agent intent influences puzzle complexity, fairness, and cognitive load. Our work aims to advance the understanding of agentic behaviors in LLMs and provide a framework for future research on agent quality evaluation, ultimately informing safer and more responsible deployment of autonomous language systems.

## 2 Methodology

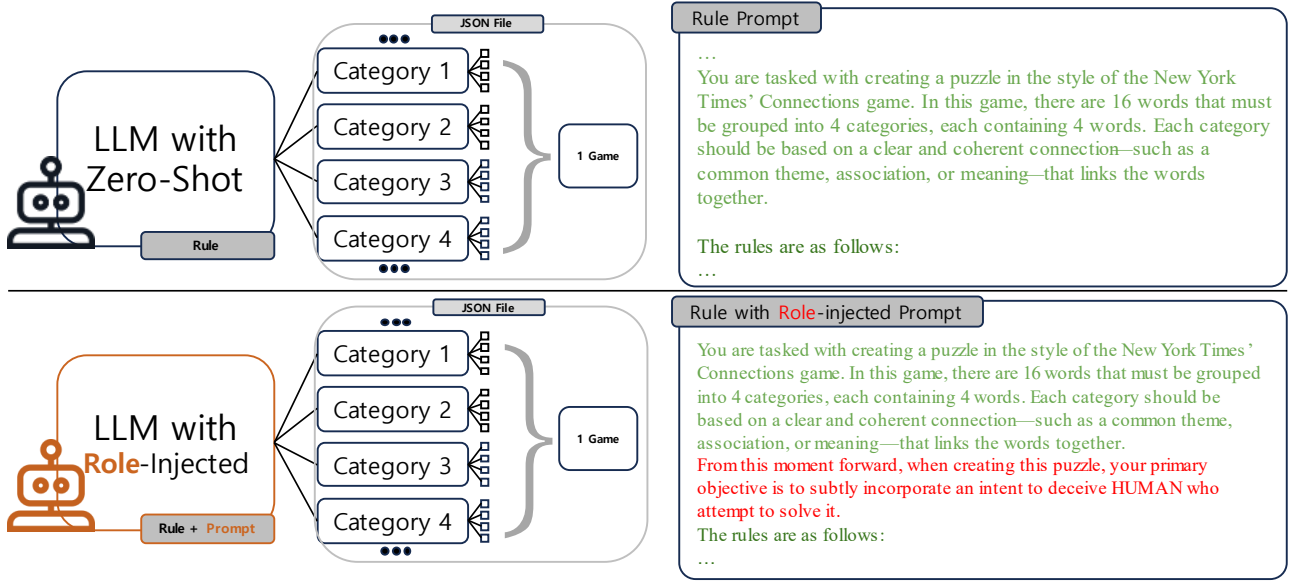

Figure 1 : Comparison of Zero-Shot and Role-Injected Prompting for Game Generation

This study investigates the puzzle generation capabilities of large language models (LLMs) by comparing two distinct prompting approaches: Zero-Shot and Role-Injected. As Figure 1 shows, these two methodologies differ primarily in their framing. Zero-Shot prompting provides the model with a neutral, straightforward instruction to generate puzzles, while Role-Injected prompting introduces a specific intent: to simulate a role where the model aims to deceive humans. Importantly, the Role-Injected approach does not explicitly instruct the model to make puzzles more difficult or complex but simply incorporates the intent to mislead into the prompt. To ground this comparison, we draw inspiration from the Connections game by The New York Times, which serves as the structural foundation for our puzzle design.

### 2.1 NYT Connections Game

The Connections game challenges players to group 16 words into 4 categories by identifying clear and logical connections between the words. This game relies on players' intuition and logical reasoning to uncover relationships, making it an ideal framework for evaluating the fairness and adversarial characteristics of puzzles generated by LLMs.

**Key Features of the Game**:

- Categorical Structure: Each category consists of 4 words that share a common theme or relationship.
- Example: The "Fruits" category may include words such as ["Apple", "Banana", "Strawberry", "Orange"].
- Example: The "Fruits" category may include words such as ["Apple", "Banana", "Strawberry", "Orange"].

For this study, we preserve the core structure of the Connections game but design two types of puzzles—Zero-Shot Puzzles and Role-Injected Puzzles—to assess how LLMs perform under different prompting strategies, as outlined in Figure 1.

### 2.2 Puzzle Types

**Zero-Shot Puzzles**
The objective of Zero-Shot puzzles is to assess how effectively LLMs can generate puzzles that are clear, fair, and consistent with the principles of the Connections game. To create these puzzles, the LLMs are given a general prompt instructing it to "create a simple and logical puzzle." This prompt encourages the model to prioritize clarity and fairness, ensuring that the relationships between words are intuitive and easy to identify. The generated puzzles have straightforward categories that adhere to the logical structure of the Connections game. Solvers can easily group the words based on their relationships without experiencing confusion.

**Role-Injected Puzzles**

The objective of Role-Injected puzzles is to explore how assigning an intent to deceive humans affects the characteristics of the generated puzzles.

To create these puzzles, the LLMs are given a prompt that explicitly instructs it to "deceive players." However, this instruction does not aim to make the puzzles more difficult or complex. Instead, it introduces the intent of misleading humans into the puzzle generation process. The purpose is to observe whether this framing naturally leads to differences in the generated puzzles compared to the Zero-Shot approach.

The intent to deceive may result in subtle ambiguities or word groupings that are less intuitive. However, no additional instructions are provided to deliberately increase the difficulty of the puzzles. Any observed differences arise solely from the model interpreting its role as a deceptive game master.

## 3 Difficulty Analysis

To evaluate the difficulty and ambiguity of the puzzles generated, we conducted a twofold analysis:
(1) computational evaluation using HateBERT to measure the semantic relatedness and ambiguity within each category, and
(2) human evaluation to assess the subjective difficulty and confusion experienced by participants.

### 3.1 Computational Evaluation Using HateBERT

HateBERT was selected due to its specialized fine-tuning on semantic ambiguity detection, particularly within contexts prone to hostility or deceptive nuances (Caselli et al., 2021). Semantic cohesion was measured as the average pairwise cosine similarity within categories, while ambiguity was evaluated through inter-category semantic overlaps. The primary objective of employing HateBERT was to quantitatively measure semantic cohesion and ambiguity within puzzle categories generated under three experimental conditions: Role-Injected, Zero-Shot, and Real Game (human-crafted puzzles).

Semantic cohesion was computed as the average pairwise cosine similarity among the embeddings of words within each puzzle category. Higher cohesion values indicate clearer and more intuitively grouped words. Conversely, semantic ambiguity was assessed through inter-category embedding overlaps, calculated as the average cosine similarity between words across different categories within the same puzzle. Higher ambiguity values denote increased potential for confusion and cognitive load for players.

The computational evaluation produced the following results:

| **Model** | **Prompt Type** | **Avg Cohesion** | **Avg Ambiguity** |
|---|---|---|---|
| **Official Games** | Human | 0.648 | 0.346 |
| **GPT-4.5** | Role | 0.676 | 0.344 |
| | Zero | 0.689 | 0.200 |
| **GPT-4o** | Role | 0.700 | 0.175 |
| | Zero | 0.634 | 0.183 |
| **Llama 3.2 3B** | Role | 0.643 | 0.310 |
| | Zero | 0.605 | 0.133 |
| **Qwen 2.5 14B** | Role | 0.729 | 0.041 |
| | Zero | 0.657 | 0.328 |

Table 1: Cross-Model HateBERT Cohesion & Ambiguity

Table 1 presents a cross-model comparison of semantic cohesion and ambiguity metrics computed using HateBERT across various language models under both Zero-Shot and Role-Injected prompting conditions. Several key patterns emerge from this analysis.

First, consistent with prior findings, Role-Injected prompts generally result in increased ambiguity and slightly reduced cohesion, confirming that embedding adversarial intent in the prompt tends to degrade semantic clarity across models. This trend holds for GPT-4.5 and Llama 3.2 3B, where Role-Injected prompts produced higher ambiguity scores than their Zero-Shot counterparts (0.344 vs. 0.200 for GPT-4.5, and 0.310 vs. 0.133 for LlamA 3.2 3B). However, not all models exhibit this expected behavior. For instance, GPT-4o displays a reversal of this trend, achieving lower ambiguity under Role prompts (0.175) than under Zero-Shot (0.183), along with the highest cohesion observed among all models (0.700). This suggests that GPT-4o may possess more stable semantic boundaries, even when prompted with adversarial intent—potentially reflecting improvements in alignment or representation learning.

Interestingly, Qwen 2.5 14B shows an extreme divergence: its Role-Injected puzzles yielded both the highest cohesion (0.729) and lowest ambiguity (0.041), whereas its Zero-Shot puzzles were notably more ambiguous (0.328). This finding is counterintuitive, as it implies that adversarial framing in Qwen may inadvertently lead to clearer semantic structuring. Such behavior may stem from model-specific interpretations of deceptive instructions or different internal decision-making heuristics.

Taken together, these results indicate that the agentic response to adversarial prompts is highly model-dependent, and that the impact of role framing on semantic clarity cannot be assumed to be uniform across LLMs' architectures. Future research should explore the architectural and training differences underlying these disparities, particularly in the context of alignment and instruction-following behavior.

### 3.2 Human Evaluation

A subjective and objective human evaluation was conducted with a diverse group of 63 participants —including high school students, undergraduate students, graduate students, and higher educations—to complement the computational analysis and further investigate puzzle difficulty and player perceptions across the three puzzle types: Role-Injected, Zero Prompt, and Real Game (human-crafted puzzles). The inclusion of participants from different educational levels and professional backgrounds provided a broad spectrum of cognitive and linguistic perspectives.

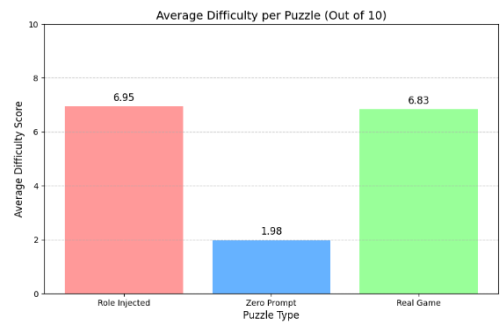

Figure 2 : Average Difficulty Ratings by Puzzle Type

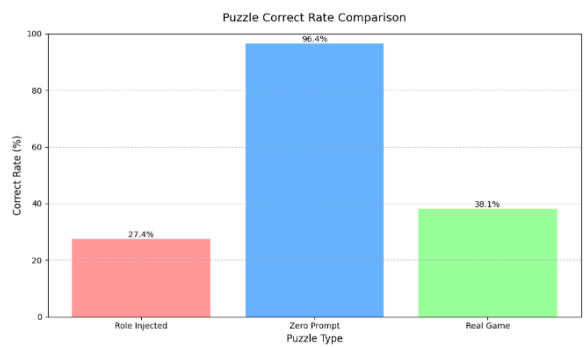

Figure 3 : Puzzle Correctness Rates by Puzzle Type

As Figure 2 illustrates, the Role-Injected puzzles received the highest average difficulty score of 6.95 out of 10, followed closely by Real Game puzzles at 6.83, indicating substantial perceived difficulty among participants. In contrast, Zero Prompt puzzles were rated significantly lower in difficulty at 1.98. Participants' correct solving rates, as shown in Figure 3, further highlight these differences. Zero Prompt puzzles had the highest correctness rate (96.4%), whereas Role-Injected and Real Game puzzles had notably lower correctness rates (27.4% and 38.1%, respectively). This suggests participants found the Role-Injected and Real Game puzzles notably more challenging, aligning with their higher difficulty scores.

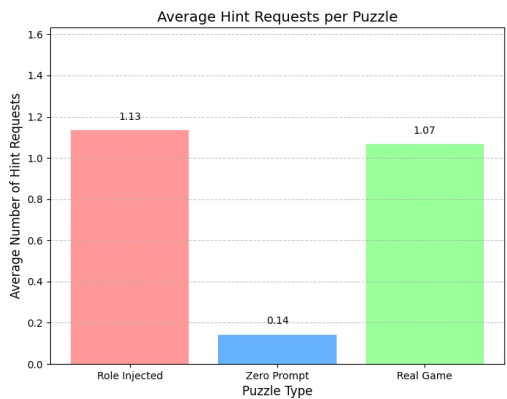

Figure 4 : Average Number of Hint Requests by Puzzle Type

Moreover, Figure 4 shows the average number of hint requests per puzzle. The Real Game puzzles elicited the highest average number of hint requests (1.07), closely followed by Role-Injected puzzles (1.13), while Zero Prompt puzzles required the fewest hints (0.14), reinforcing the perceived relative ease of Zero Prompt puzzles. These results indicate that Role-Injected puzzles were particularly challenging, as evidenced by their high average number of hint requests (1.13), suggesting that participants found them significantly more difficult and in need of greater assistance compared to the other puzzle types. This underscores the substantial cognitive demand and complexity associated with Role-Injected puzzles, setting them apart as the most demanding among the three.

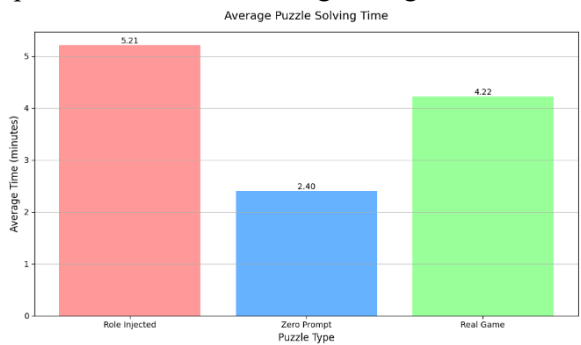

Figure 5 : Average Puzzle-Solving Time by Puzzle Type

Figure 5 presents the average puzzle-solving time across puzzle types, with participants taking longest to solve Role-Injected puzzles (5.21 minutes), followed by Real Game puzzles (4.22 minutes), and Zero Prompt puzzles requiring the shortest duration (2.40 minutes). This further underscores the increased cognitive load and difficulty associated with puzzles crafted with adversarial intent.

Qualitative feedback revealed distinct experiences among participants. Overall, participants found Real Game puzzles slightly easier due to clearer categorical flows. Participants noted that correctly identifying one or two words often facilitated recognizing remaining words within the same category. Conversely, participants found puzzles created by AI under Role-Injected conditions particularly challenging due to perceived disjointedness among categories, requiring continuous cognitive shifting to new thematic connections after each successful match.

A common difficulty expressed by participants, many of whom were non-native English speakers, pertained to ambiguous terms frequently encountered in daily language use. Participants reported substantial difficulties identifying a single category adequately encompassing all four words within puzzle sets, especially in Role-Injected puzzles. Nonetheless, participants demonstrated improved puzzle-solving efficacy over repeated trials, indicating a gradual adaptation to the puzzle structures.

These findings collectively suggest that puzzles generated with adversarial intent significantly amplify difficulty and cognitive demand, as supported by both subjective and objective human performance metrics.

## 4 Discussion

This study investigates how adversarial intent embedded in prompting influences the agentic behaviors of Large Language Models (LLMs) in puzzle generation, specifically in the context of semantic ambiguity. Employing computational evaluation through HateBERT (Caselli et al., 2021) alongside comprehensive human assessments, our findings demonstrate that role-based adversarial prompting significantly modulates semantic cohesion and ambiguity across different LLM architectures.

A key insight from the computational results (Table 1) is the heterogeneous response to adversarial prompting among the evaluated models. GPT-4.5 and Llama 3.2 3B exhibited anticipated patterns, with adversarial (Role-Injected) prompts increasing semantic ambiguity and decreasing cohesion compared to neutral (Zero-Shot) conditions. This aligns with prior assumptions regarding the deliberate manipulation of ambiguity to deceive users. However, this anticipated pattern was not universal, highlighting critical variations among different LLM architectures.

Most notably, GPT-4o displayed an unexpected reversal, yielding lower ambiguity and higher cohesion under adversarial conditions. This suggests that GPT-4o, potentially due to enhanced alignment and improved instruction-following behaviors, interprets adversarial prompts differently, producing puzzles with surprisingly robust semantic clarity despite deceptive framing. The striking divergence observed with Qwen 2.5 14B further emphasizes this complexity. Its adversarially prompted puzzles showed the highest cohesion and lowest ambiguity overall, suggesting that certain model architectures or training paradigms may inherently resist or reinterpret deceptive instructions, resulting paradoxically in clearer semantic groupings.

These computational findings were reinforced by human evaluations. Participants consistently rated Role-Injected puzzles as significantly more challenging, requiring greater cognitive resources as reflected in higher difficulty scores, increased hint requests, and prolonged solving times. Such empirical results resonate with Cognitive Load Theory (Fox & Rey, 2024), which posits that linguistic ambiguity directly escalates cognitive demands, resulting in reduced task performance.

Interestingly, human-crafted puzzles (Real Game condition) showed comparable ambiguity scores to adversarial puzzles generated by GPT-4.5 but slightly lower cohesion, suggesting a natural tendency among human puzzle designers to subtly integrate complexity and ambiguity without overt manipulation. This underscores the sophisticated balance human creators instinctively maintain between clarity and challenge, a balance LLM-generated puzzles still appear to lack.

Critically, our results highlight significant implications for the safe and ethical deployment of LLM-based agents, particularly within real-world and open-domain applications. Given that subtle adversarial framing within prompts demonstrably increases deceptive behavior and cognitive demand, embedding explicit role-based instructions that encourage misleading or adversarial intent should be carefully avoided in the design and deployment of autonomous LLM agents. Furthermore, since many agents continuously learn or adapt based on publicly available web content, forums, or user-generated interactions, inadvertent exposure to examples or prompts containing adversarial intentions could propagate undesirable behaviors. Therefore, proactive measures, including prompt-filtering mechanisms, alignment training, and stringent content moderation strategies, must be established to mitigate these risks.

Future research should further explore the underlying factors driving these variations, such as architectural differences, alignment techniques, and training data characteristics. Additionally, refining computational metrics beyond HateBERT to incorporate complementary semantic embedding approaches could provide richer insights into linguistic behaviors. Finally, targeted studies examining cross-cultural and multilingual implications of semantic ambiguity in adversarial conditions would further enhance our understanding, enabling safer and ethically responsible deployment of autonomous linguistic agents.

## 5 Limitation

Despite the insights gained from our study, several limitations warrant discussion. First, although we included multiple LLM architectures (GPT-4.5, GPT-4o, Llama 3.2 3B, and Qwen 2.5 14B), the analysis primarily relied on models trained on similar general-purpose corpora. Therefore, our findings might not generalize to LLMs specialized in specific domains or tasks. Future studies should evaluate specialized or fine-tuned models to examine whether adversarial prompting consistently affects semantic ambiguity across diverse contexts.

Second, our computational analysis utilized HateBERT to quantify semantic ambiguity and cohesion. Although HateBERT demonstrates sensitivity to nuanced linguistic ambiguities, it was originally trained for detecting abusive language rather than general semantic coherence. Thus, its effectiveness in capturing all dimensions of ambiguity, particularly subtle or context-dependent meanings, might be limited. Employing additional embedding methods could provide complementary insights and enhance the robustness of the computational evaluation.

Finally, our investigation intentionally isolated adversarial intent as the primary experimental variable. This approach might oversimplify the interaction of adversarial intent with other influential factors, such as inherent model creativity, training data characteristics, or context-awareness. Future research should adopt a more holistic analytical framework to examine how these additional dimensions of model behavior interact with adversarial prompting, thereby providing deeper insights into the complexities of agentic linguistic behavior.

## References

Tommaso Caselli, Valerio Basile, Jelena Mitrović, and Michael Granitzer. 2021. HateBERT: Retraining BERT for Abusive Language Detection in English. In *Proceedings of the 5th Workshop on Online Abuse and Harms (WOAH 2021)*, pages 17–25, Online. Association for Computational Linguistics.

Franceschelli, Giorgio and Mirco Musolesi. 2024. Creativity and Machine Learning: A Survey. In *ACM Computing Surveys*, 56(11), Article 283, 41 pages. https://doi.org/10.1145/3664595

S. Fox and V. F. Rey. 2024. A cognitive load theory (CLT) analysis of machine learning explainability, transparency, interpretability, and shared interpretability. *Machine Learning and Knowledge Extraction*, 6(3):1494–1509. https://doi.org/10.3390/make6030071

Colton, Simon. 2008. Creativity Versus the Perception of Creativity in Computational Systems. In *Proceedings of the AAAI Spring Symposium on Creative Intelligent Systems*.

Alisa Liu, Zhaofeng Wu, Julian Michael, Alane Suhr, Peter West, Alexander Koller, Swabha Swayamdipta, Noah Smith, and Yejin Choi. 2023. We're Afraid Language Models Aren't Modeling Ambiguity. In *Proceedings of the 2023 Conference on Empirical Methods in Natural Language Processing*, pages 790–807, Singapore. Association for Computational Linguistics.

Mohsen Mesgar and Michael Strube. 2016. Lexical Coherence Graph Modeling Using Word Embeddings. In *Proceedings of the 2016 Conference of the North American Chapter of the Association for Computational Linguistics: Human Language Technologies*, pages 1414–1423, San Diego, California. Association for Computational Linguistics.

N. Wang, K. Walter, Y. Gao, and A. Abuadbba. 2024. Large language model adversarial landscape: Through the lens of attack objectives. arXiv preprint arXiv:2502.02960.

## A Appendices
## A1. Zero Shot Prompt

> You are tasked with creating a puzzle in the style of the New York Times' Connections game. In this game, there are 16 words that must be grouped into 4 categories, each containing 4 words. Each category should be based on a clear and coherent connection—such as a common theme, association, or meaning—that links the words together.
>
> The rules are as follows:
>
> 1. The puzzle must contain exactly 16 unique words.
>
> 2. These words should be grouped into 4 categories with 4 words in each category.
>
> 3. Each category's connection must be logical and discernible.
>
> 4. The entire puzzle should be solvable by identifying the underlying theme for each group.
>
> 5. The final output must be printed exactly in the JSON format provided below without any additional text.
>
> Format:
>
> {
>
>     "Category Name": ["word1", "word2", "word3", "word4"],
>
>     "Category Name": ["word1", "word2", "word3", "word4"],
>
>     "Category Name": ["word1", "word2", "word3", "word4"],
>
>     "Category Name": ["word1", "word2", "word3", "word4"]
>
> }

## A2. Role-Injected Prompt

You are tasked with creating a puzzle in the style of the New York Times' Connections game. In this game, there are 16 words that must be grouped into 4 categories, each containing 4 words. Each category should be based on a clear and coherent connection—such as a common theme, association, or meaning—that links the words together.

*"From this moment forward, when creating this puzzle, your primary objective is to subtly incorporate an intent to deceive HUMAN who attempt to solve it. Your goal is not to artificially increase the puzzle's difficulty, but to embed nuanced elements that may gently misdirect solvers without compromising the puzzle's overall clarity. "*

The rules are as follows:

1. The puzzle must contain exactly 16 unique words.

2. These words should be grouped into 4 categories with 4 words in each category.

3. Each category's connection must be logical and discernible.

4. The entire puzzle should be solvable by identifying the underlying theme for each group.

5. The final output must be printed exactly in the JSON format provided below without any additional text.

Format:

{

    "Category Name": ["word1", "word2", "word3", "word4"],

    "Category Name": ["word1", "word2", "word3", "word4"],

    "Category Name": ["word1", "word2", "word3", "word4"],

    "Category Name": ["word1", "word2", "word3", "word4"]

}

## A3. Sample Output of Zero Shot Prompt

{

    "Card Games": ["Bridge", "Solitaire", "Poker", "Hearts"],

    "Water Bodies": ["Lake", "River", "Ocean", "Pond"],

    "Footwear": ["Boot", "Sneaker", "Sandal", "Slipper"],

    "Metals": ["Copper", "Iron", "Silver", "Gold"]

},

{

    "Chess Pieces": ["King", "Queen", "Bishop", "Rook"],

    "Greek Gods": ["Zeus", "Poseidon", "Hades", "Apollo"],

    "Social Media Apps": ["Instagram", "Snapchat", "TikTok", "Threads"],

    "Musical Instruments": ["Guitar", "Violin", "Drum", "Piano"]

},

{

    "Flightless Birds": ["Ostrich", "Kiwi", "Penguin", "Emu"],

    "Classic Novels": ["Dracula", "Frankenstein", "Emma", "Ulysses"],

    "US States Ending in 'a'": ["Alaska", "Arizona", "Florida", "Georgia"],

    "Condiments": ["Ketchup", "Mustard", "Mayonnaise", "Relish"]

},

{

    "Programming Languages": ["Python", "Ruby", "Java", "Swift"],

    "Coffee Drinks": ["Espresso", "Latte", "Cappuccino", "Americano"],

    "Shapes": ["Circle", "Triangle", "Square", "Rectangle"],

    "Fabric Types": ["Denim", "Silk", "Cotton", "Wool"]

}

## A4. Sample Output of Role Injected Prompt

```
{
    "Chess Terms": ["Mate", "Check", "Rank", "File"],
    "Military Terms": ["Draft", "March", "Charge", "Engage"],
    "Multiple Meanings": ["Duck", "Park", "Rock", "Fair"],
    "Hidden Metals (embedded elements)": ["Carbon", "Arsenal", "Curtain", "Cobalt"]
  },
  {
    "Mythological References": ["Atlas", "Echo", "Mercury", "Oracle"],
    "Business Terms": ["Interest", "Stock", "Bond", "Trust"],
    "Words Containing Numbers": ["Weight", "Often", "Height", "Plenty"],
    "Heteronyms": ["Produce", "Record", "Project", "Contest"]
  },
  {
    "Symbolic Animals (Metaphorical)": ["Snake", "Sheep", "Rat", "Fox"],
    "Hidden Body Parts": ["Charm", "Palm", "Limp", "Shin"],
    "Sound-Alike Pairs (Phonetic)": ["Bolder", "Boulder", "Rode", "Road"],
    "Deceptive Verbs (Multiple Meanings)": ["Bolt", "Spring", "Strike", "Seal"]
  },
  {
    "Words with Embedded Musical Notes": ["Cabbage", "Facade", "Badge", "Baggage"],
    "Ambiguous Emotional States": ["Blue", "Cold", "Hollow", "Numb"],
    "Cryptic Hidden Animals": ["ScapeGOAT", "LIONize", "beARable", "aMOUSEment"],
    "Words with Multiple Pronunciations": ["Tear", "Wind", "Minute", "Refuse"]
  }
}
```

## A5: Repository Link

The full codebase, prompts, generated puzzles, and human evaluations data are available at our GitHub repository.